%% file: main.tex
\pdfoutput=1

\documentclass[11pt]{article}

\usepackage{acl}

\usepackage{times}
\usepackage{latexsym}
\usepackage{booktabs}
\usepackage{multirow}
\usepackage{graphicx}

\usepackage[T1]{fontenc}

\usepackage[utf8]{inputenc}

\usepackage{microtype}

\definecolor{forestgreen}{RGB}{34,139,34}

\title{AnnIE: An Annotation Platform for Constructing Complete \\ Open Information Extraction Benchmark}

\author{Niklas Friedrich$^{1}$, Kiril Gashteovski$^2$, Mingying Yu$^{1,2}$, Bhushan Kotnis$^{2}$, 
        \\ {\bf Carolin Lawrence$^{2}$,} {\bf Mathias Niepert$^{2,3}$,} {\bf Goran Glava\v{s}$^{1,4}$} \\
        $^1$ University of Mannheim, Mannheim, Germany \\ 
        $^2$ NEC Laboratories Europe, Heidelberg, Germany \\ 
        $^3$ University of Stuttgart, Stuttgart, Germany, $^4$ LMU Munich, Munich, Germany \\ 
        \texttt{\{nfriedri,minyu,gglavas\}@mail.uni-mannheim.de} \\ 
        \texttt{\{firstname.lastname\}@neclab.eu}
        }
        
\begin{document}
\maketitle
\begin{abstract}


Open Information Extraction (OIE) is the task of extracting facts from sentences in the form of relations and their corresponding arguments in schema-free manner. Intrinsic performance of OIE systems is difficult to measure due to the \textit{incompleteness} of existing OIE benchmarks:~ground truth extractions do not group all acceptable surface realizations of the same fact that can be extracted from a sentence. 
To measure performance of OIE systems more realistically, it is necessary to manually annotate \textit{complete facts} (i.e., clusters of all acceptable surface realizations of the same fact) from input sentences. 
We propose AnnIE: an interactive annotation platform that facilitates such challenging annotation tasks and supports creation of \textit{complete fact-oriented OIE evaluation benchmarks}. 
AnnIE is modular and flexible in order to support different use case scenarios 
(i.e., benchmarks covering different types of facts) and different languages. 
We use AnnIE to build two complete OIE benchmarks: one with verb-mediated facts and another with facts encompassing named entities.    
We evaluate several OIE systems on our complete benchmarks created with AnnIE. 
We publicly release AnnIE under non-restrictive license.\footnote{\url{https://github.com/nfriedri/annie-annotation-platform}}

\end{abstract}

\input{1-intro}
\input{2-rel-work}
\input{3-BenchIE}
\input{4-platform-overview}

\input{5-experiments}

\input{6-conclusions}

\bibliography{main}
\bibliographystyle{acl_natbib}

\clearpage

\appendix
\input{appendix}

\end{document}

%% file: 1-intro.tex
\section{Introduction}


Open Information Extraction (OIE) is the task of extracting relations and their arguments from natural language text 
in schema-free 
manner \cite{banko2007open}. Consider the input sentence \emph{"Edmund Barton, who was born in Australia, 
was a judge"}. Without the use of a pre-specified schema, an OIE system should extract the
triples \emph{("Edmund Barton"; "was born in"; "Australia")} and \emph{("Edmund Barton"; "was"; "judge")}.
The output of OIE systems is used in many downstream tasks, including open link prediction \cite{broscheit2020}, 
automated knowledge base construction \cite{gashteovski2020aligning}, question answering \cite{khot2017answering} and text summarization \cite{xu2021generating}.

Intrinsic evaluation of OIE systems is done either manually 
\cite{mausam2012open,pal2016demonyms} or with the use of evaluation benchmarks 
\cite{Stanovsky2016EMNLP,bhardwaj2019carb}. While manual evaluations are usually of higher quality, they are
expensive and time consuming. 
Automated benchmark evaluations are faster and more economic than manual OIE evaluations 
\cite{hohenecker2020systematic}, but are less reliable than human judgments of extraction correctness \cite{zhan2020span}, because they are based on approximate token-level matching of system extractions against ground truth extractions.
The main shortcoming of existing OIE benchmarks is their \textit{incompleteness}: they do not exhaustively list all acceptable surface realizations of the same piece of information (i.e., same fact) and, because of this, resort to unreliable scoring functions based on token-level matching between system and gold extractions \cite{schneider2017relvis}.

Obtaining complete manual OIE annotations is, however, very difficult and time-consuming. 
Annotating a \textit{complete} OIE benchmark requires human annotators to write \emph{all} 
possible combinations of extractions expressing the same fact  (i.e., exhaustively list all acceptable surface realizations of the same fact; see Section~\ref{sec:benchie}). To facilitate and speed up this process, we introduce AnnIE, a dedicated annotation tool for constructing complete fact-oriented OIE benchmark.   
AnnIE facilitates the annotation process by (1) highlighting the tokens of interest (e.g., for verb-mediated extractions, it highlights the verbs, which are candidates for head words of predicates); 
(2) providing web-based interface for annotating triples and grouping them into \textit{fact synsets}, i.e., groups of informationally equivalent extractions (Section~\ref{sec:benchie}). 
To the best of our knowledge, AnnIE is the first publicly-available annotation platform for constructing OIE benchmarks. 


We showcase AnnIE\footnote{Video demo: \url{https://youtu.be/2wn75U8Lc5w}} by creating two complete fact-based OIE benchmarks: (1) benchmark of verb-mediated facts on English, German, Chinese, Galician, Arabic and Japanese, making this gold data the first such OIE resource on languages other than English; (2) benchmark for facts associating named entities (for English only). We then benchmark several state-of-the-art OIE systems on these fact-based benchmarks and demonstrate that they are significantly less effective than indicated by existing OIE benchmarks that use token-level scoring. We hope that AnnIE motivates the creation of  many more fact-based (as opposed to token-level) OIE evaluation benchmarks.





%% file: 2-rel-work.tex
\section{Related Work}

\subsection{Evaluation of OIE Systems}
OIE systems are evaluated either manually
\cite{mausam2012open,pal2016demonyms,Gashteovski2019OPIECAO}, w.r.t.~a downstream task \cite{mausam2016open,lin2020kbpearl}, or 
with the use of evaluation benchmarks \cite{Stanovsky2016EMNLP,bhardwaj2019carb}. Manual evaluations are usually of higher quality because they are performed by one or more expert annotators \cite{del2013clausie}. They are, however, expensive and time consuming, which makes the development of OIE systems very slow. 
On the other hand, downstream evaluation of OIE systems is faster, but provides insights only about their performance w.r.t.~particular tasks and does not provide 
insights on the intrinsic (i.e., task-agnostic) correctness of the extractions. Finally, using evaluation benchmarks is both task-agnostic and fast, though current benchmarks might contain noise 
\cite{zhan2020span}. Moreover, current benchmarks suffer from incompleteness; i.e., they are not designed in a
manner that aims to contain all possible extractions from an input sentence. Therefore, they rely on lenient 
token-overlap based evaluation, which could result in misleading results \cite{lechelle2019wire57}. To address
this, we move away from such token-based evaluations and move towards fact-based evaluation
(Section~\ref{sec:benchie}).

\subsection{Annotation Tools}

To facilitate the annotation process of NLP tasks, many interactive annotation tools have been designed. 
Such work covers tasks like sequence labelling \cite{lin2019alpacatag,lee2020lean}, coreference resolution 
\cite{bornstein-etal-2020-corefi} and treebank projection across languages
\cite{akbik2017projector}. 
For annotating OIE extractions, however, there are no publicly available tools. 
The two commonly used benchmarks---OIE2016 \cite{Stanovsky2016EMNLP} and CaRB \cite{bhardwaj2019carb}---only provide annotated data and no dedicated annotation tool. 
OIE2016 uses a dataset from a similar task (QA-SRL), which is then automatically
ported to OIE. 
This approach does not require an annotation tool, but the quality of the benchmark (i.e., ground truth extractions) decreases due to the automatic label projection \cite{zhan2020span}.
CaRB addresses this issue 
by sampling from the same input sentences used by OIE2016, and then crowdsourcing manual extractions. However, their annotation OIE interface has four major limitations: (1) it cannot be used to create complete fact-based OIE benchmarks (Section~\ref{sec:benchie}), i.e., it does not allow for different extractions (e.g., triples) that correspond to the same fact; this leads to incomplete annotations and unreliablly lenient token-overlap-based evaluation measures; (2) it focuses only on one type of OIE (verb-mediated extractions); (3) it is not publicly available; (4) it does not support annotations for languages other than English.



%% file: 3-BenchIE.tex
\section{Fact-Based OIE Evaluation}
\label{sec:benchie}

\begin{table*}
    \centering
    \footnotesize
        \begin{tabular}{rrlccc}  
            \toprule
            \multicolumn{6}{l}{\textbf{Input sentence:} \emph{"Sen.~Mitchell is confident he has sufficient votes to block such a measure with procedural actions."}} \\
            \multicolumn{6}{l}{\textbf{CaRB gold extraction:} \emph{("Sen.~Mitchell"; "is confident he has"; "sufficient votes to block ...procedural actions")}} \\
            \cmidrule(r){1-6}
            & \multicolumn{2}{c}{Input OIE extraction}    & \multicolumn{2}{c}{CaRB (P / R)} & Fact-based \\
            $t_1$ & \emph{("Sen.~Mitchell"; "is confident he has";} & \emph{ "sufficient")} & 1.00 & 0.44 & 0 \\ 
            $t_2$ & \emph{("Sen.~Mitchell"; "is confident he has";} & \emph{ "sufficient actions")} & 1.00 & 0.50 & 0 \\
            $t_3$ & \emph{("Sen.~Mitchell"; "is confident he has";} & \emph{ "sufficient procedural actions")} & 1.00 & 0.56 & 0 \\
            \cmidrule(r){1-6}
            $t_4$ & \emph{("Sen.~Mitchell"; "is confident he has";} & \emph{ "sufficient votes")} & 1.00 & 0.50 & 1 \\ 
            \bottomrule
        \end{tabular}
    
    \caption{Difference in scores between CaRB and fact-based evaluation. For the input sentence, CaRB provides only one extraction which covers all the words in the sentence. 
    Then, for each input OIE extraction (from $t_1$ to $t_4$) it calculates token-wise precision and recall scores w.r.t.~the golden annotation. Fact-based evaluation (with all acceptable extractions of the fact exhaustively listed) allows for exact matching against OIE extractions.} 
    \label{tab:carb-benchie-extractions2}
\end{table*}

Due to their incompleteness, previous benchmarks lack clarity about whether an extraction indeed represents a correct fact or not. In particular, given a system extraction, they do not assign a binary score (correct/incorrect), but rather calculate per-slot token 
overlap scores. Consider, for example, the input sentence from Table~\ref{tab:carb-benchie-extractions2} and the scores that the recent OIE benchmark CaRB \cite{bhardwaj2019carb} assigns to extractions $t_1$ to $t_3$. 
Because all tokens for each slot for $t_1-t_3$ are also present in the gold extraction, CaRB credits these extractions with a perfect precision score, even though the extractions clearly state incorrect facts. In similar vein, the CaRB recall score of the extraction $t_4$ is lower than the recall score of $t_3$, even though $t_4$ captures the correct core fact and $t_3$ does not. 

To address these issues, we propose moving away from such lenient token-overlap scoring and going towards 
fact-level exact matching. To this end, we propose an evaluation framework, dubbed BenchIE \cite{gashteovski2021benchie}, for OIE evaluation based on facts, not 
tokens. 
Here, the annotator is instructed to \emph{exhaustively} list all possible surface realizations of the same fact, allowing for a binary judgment (correct/incorrect) of correctness of each extraction (it either exactly matches some of the acceptable gold realizations of some fact or it does not match any). 
The example in Table~\ref{tab:benchie-example} illustrates the concept of a \textit{fact synset}: a collection of all acceptable extractions for the same fact (i.e., same piece of knowledge). 

Because benchmarks based on fact synsets are supposed to be complete, a system OIE extraction is considered correct if and only if it \emph{exactly} 
matches any of the gold extractions from any of the fact synsets. The number of \textit{true positives (TPs)} is 
the number of fact synsets (i.e., different facts) ``covered'' by at least one system extraction. This way, a 
system that extracts $N$ different triples of the same fact, will be rewarded only once for the correct extraction 
of the fact. False negatives (FNs) are then fact synsets not covered by any of the system 
extractions. Finally, each system extraction that does not exactly match any gold triple (from any synset) is 
counted as a false positive (FP). We then compute \textit{Precision}, \textit{Recall}, and $F_1$ score from TP, FP, and FN in the standard fashion. For more details on the evaluation framework, see \cite{gashteovski2021benchie}.

\begin{table*}
    \footnotesize
    \def\arraystretch{0.8}
        \begin{tabular}{rrrr}  
            \toprule
            \multicolumn{4}{l}{\textbf{Input sentence:} \emph{"Sen.~Mitchell is confident he has sufficient votes to block such a measure with procedural actions."}} \\
            \midrule
            $f_1$ & \emph{("Sen.~Mitchell" | "he";} & \emph{"is";} & \emph{"confident [he has sufficient ... actions]")} \\
            \midrule
            $f_2$ & \emph{("Sen.~Mitchell" | "he";} & \emph{"is confident he has";} & \emph{"sufficient votes")} \\
                  & \emph{("Sen.~Mitchell" | "he";} & \emph{"is confident he has";} & \emph{"suff.~votes to block [such] [a] measure")} \\
            \midrule
            $f_3$ & \emph{("Sen.~Mitchell" | "he";} & \emph{"is conf.~he has sufficient votes to block"} & \emph{"[such] [a] measure")} \\
                  & \emph{("Sen.~Mitchell" | "he";} & \emph{"is confident he has ... to block [such]";} & \emph{"[a] measure")} \\
                  & \emph{("Sen.~Mitchell" | "he";} & \emph{"is confident he has ... to block [such] [a]";} & \emph{"measure")} \\
            \midrule
            $f_4$ & \emph{("Sen.~Mitchell" | "he";} & \emph{"is conf.~he has ... [such] [a] measure with";} & \emph{"procedural actions")} \\
                  & \emph{("Sen.~Mitchell" | "he";} & \emph{"is confident he has ... [such] [a] measure";} & \emph{"with procedural actions")} \\
            \bottomrule
        \end{tabular}
        \vspace{-0.5em}
         \caption{Example sentence with four \textit{fact synsets} ($f_1$--$f_4$). We account for entity coreference and accept both \emph{"Sen.~Mitchell"} and \emph{"he"} as subjects: the delimiter ``|'' is a shorthand notation for different extractions. In the same vein, square brackets ([]) are a shorthand notation for multiple extractions: triples both with and without the expression(s) in the brackets are considered correct.} 
    \label{tab:benchie-example}
    \vspace{-1em}
\end{table*}





%% file: 4-platform-overview.tex
\section{AnnIE: Platform Description}

AnnIE is a web-based platform that facilitates manual annotations of fact-based OIE benchmarks. In this section, we discuss: (1) the functionality of highlighting tokens of interest; (2) how AnnIE facilitates creation of complete fact-based OIE benchmarks; (3) AnnIE's software architecture; and (4) AnnIE's web interface and its 
multilingual support.

\subsection{Tokens of Interest}

One of the key functionalities of AnnIE is its ability to highlight \emph{tokens of interest} -- tokens  
that commonly constitute parts of extractions of interest. For example, most OIE systems focus on extracting 
verb-mediated triples \cite{angeli2015leveraging,kolluru2020imojie}. In such case, 
\textit{verbs} clearly represent tokens of interest and are candidates for head words of fact predicates. 
Other example of tokens of interest may be named entities, which could be useful for extracting information
from domain-specific text. There has been prior work on extracting open information from specific 
domains, including the biomedical \cite{wang2018open}, legal \cite{siragusa2018relating} and 
scientific domain \cite{lauscher2019minscie}. In this work, it is important to extract open relations between named entities. Accordingly, highlighting mentions of named entities then facilitates manual extraction of the type of facts that the benchmark is supposed to cover (i.e., relations between named entities). 
AnnIE allows the user to define a custom function that yields the tokens of interest from the input sentence and then highlights these tokens for the annotator with a background color (Figure~\ref{fig:highlight}). 

To further facilitate the manual annotations, future versions of AnnIE could also include recommendations for whole OIE triples (e.g., by recommending high-confidence extractions from already existing OIE systems) or for slots (e.g., given a subject, recommend a potential relation; or given a relation, recommend candidates for the arguments). We leave such improvements for future work.



\subsection{Annotating Fact Synsets}
\begin{figure}[h]
    \includegraphics[scale=0.39]{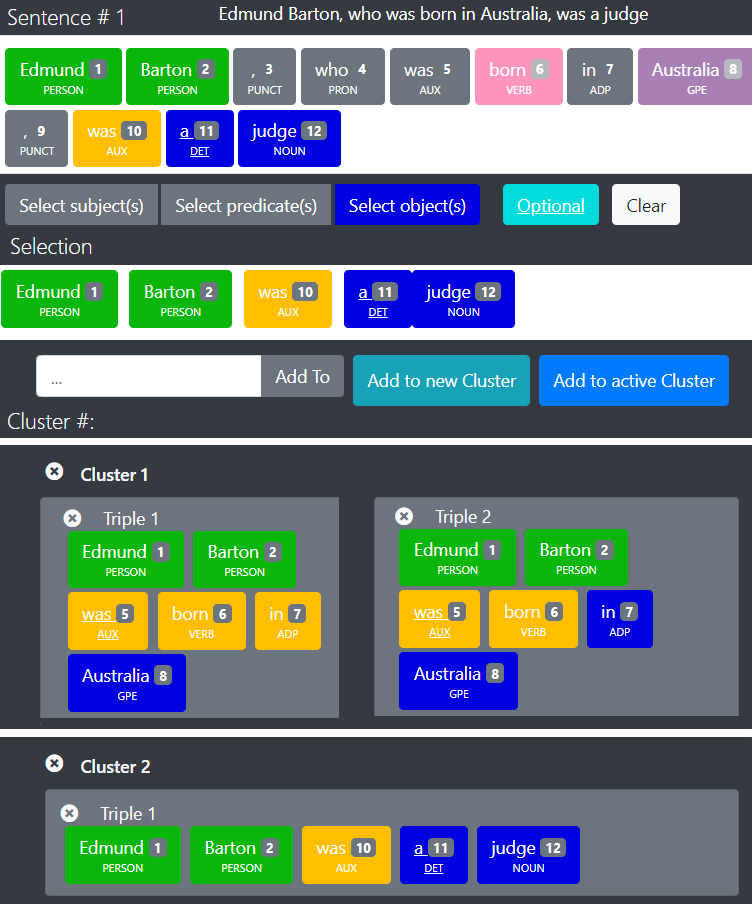}
    \caption{Manual labeling of OIE triples. The user selects tokens from the tokenized input sentence and 
    places them into the correct slot: \textcolor{forestgreen}{subject (green)}, \textcolor{orange}{predicate (yellow)} or \textcolor{blue}{object (blue)}. Then, the user adds the extracted triple either to an active fact cluster (i.e., fact synset) or to a new one. The user can also select which tokens are optional by clicking the "Optional" button on an active token selection. For larger version of the same figure, see Figure \ref{fig:fact-label-large} in Appendix \ref{app:gui}.}
    \label{fig:fact-label}
    \vspace{-1em}
\end{figure}

Given a sentence with highlighted tokens of interest, the annotator can start constructing fact synsets. Fact synsets are clusters of fact-equivalent extractions. AnnIE currently supports only the annotation of triples: for each extraction/triple the user first selects which slot she wants to annotate (subject, predicate, or object) and then selects the tokens that constitute that slot. Each token (part of one of the three slots) can additionally be marked as ``optional'', which means that the tool will create variants of that extraction both with and without those tokens.  
Once a triple is fully denoted (i.e., tokens for all three slots selected), the annotator chooses whether (1) the triple is a different variant of an already existing fact (i.e., existing fact synset), in which case the triple is added to an existing cluster or (2) a first variant of a new fact, in which case a new cluster (i.e., fact synset) is created and the triple added to it. To facilitate this decision, the annotator can at any time review the already existing fact synsets. Figure~\ref{fig:fact-label} shows the interface for creating triples and adding them to fact synsets.



%


\subsection{Platform Architecture}

\begin{figure}
    \includegraphics[scale=0.36]{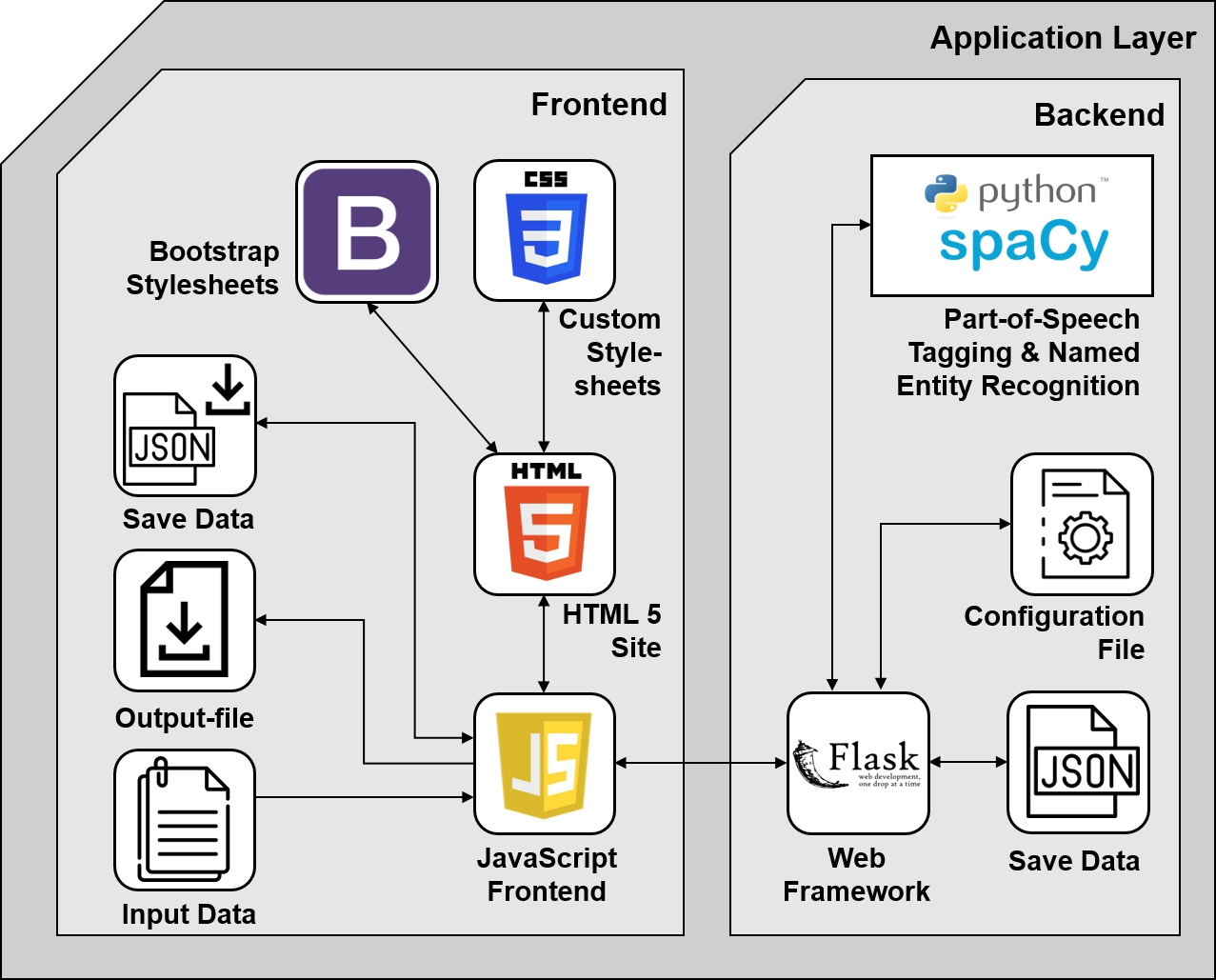}
    \caption{Software architecture of AnnIE.}
    \label{fig:annie_architecture}
    \vspace{-1em}
\end{figure}

AnnIE is a simple local executable web application (Figure~\ref{fig:annie_architecture}) that consists of a backend layer and a frontend layer. It starts a local server that provides a user interface accessible from any browser that supports JavaScript.  

\vspace{1.2mm}
\noindent\textbf{Backend.} AnnIE's backend server is based on \texttt{Flask}\footnote{\url{https://github.com/pallets/flask}}, a popular light-weight Python web framework. We implemented HTTP endpoints for receiving requests and sending responses. Additionally, \texttt{Flask} hosts the frontend (HTML and CSS) files as a web server. The \texttt{Flask} server interacts with 
(1) the NLP library \texttt{SpaCy}\footnote{\url{https://spacy.io/}} (which we employ for POS-tagging and NER, in service of highlighting tokens of interest); (2) a configuration file; (3) data files on the local hard drive and (4) the frontend (i.e., the web interface).
AnnIE's backend is highly modularized, so that any component may easily be further customized or replaced with a different module. For example, the \texttt{SpaCy}-based NLP module (for POS-tagging and NER) can easily be replaced with any other NLP 
toolkit, e.g., \texttt{Stanza}\footnote{\url{https://stanfordnlp.github.io/stanza/}} \cite{qi2020stanza}. AnnIE is also easily customizable through a configuration 
file, where the user may specify the types of tokens to be highlighted or select colors for highlighting. The I/O module expects the input (a collection of sentences for OIE annotation) to be in JSON format and saves the annotated fact synsets in JSON as well. 



\vspace{1.2mm}
\noindent\textbf{Frontend.} The application frontend is implemented in JavaScript and based on the \texttt{Bootstrap} library and custom CSS stylesheets. 
We adopt model-view-controller (MVC) architecture for the frontend: it uses a data structure (i.e., \textit{model}) capturing the entire annotation process (i.e., information about the annotation, loaded text file, current triple in annotation, etc.; you can find more details in the Appendix, Section~\ref{app-subsec:data-model}). Based on the current state of the model, the frontend renders the interface (i.e., view) by enabling and disabling particular annotation functionality. The controller connects the two: it renders the view based on the current state of the model.
%
%
We implemented additional I/O scripts that complement the core functionality of the main controller.
These scripts handle the formatting of the output file as well as the loading of the data from the input files.
Saving or loading data is customizable: one merely needs to overwrite the \texttt{load()} and \texttt{save()} methods of the controller.

\begin{figure}
    \includegraphics[scale=0.37]{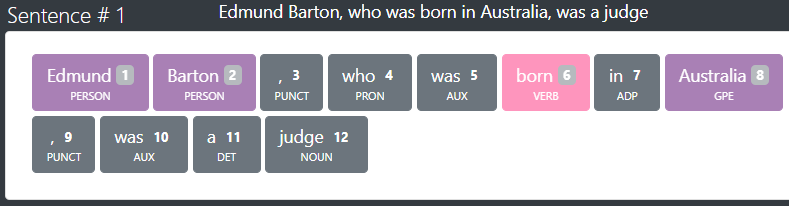}
    \caption{Highlighting tokens of interest. In this example, tokens of interest are verbs and named entities.}
    \label{fig:highlight}
    \vspace{-1em}
\end{figure}

\vspace{1.2mm}\noindent\textbf{Data Flow.} Upon selection of the input file, the data is sent from the frontend to the backend via an HTTP request. In the backend, the sentence is then tokenized, POS-tagged, and processed for named entities with the NLP module (we rely on \texttt{SpaCy}). 
The tokenized and labeled sentence is then sent as a JSON object to the frontend, where each token is displayed as one button (Figure~\ref{fig:highlight}). The default version of the tool allows the user to choose (in the configuration file) between four coloring schemes for highlighting token buttons. 


\begin{figure}
\centering
    \includegraphics[scale=0.5]{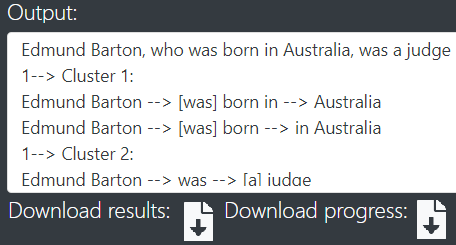}
    \caption{Human-readable representation of the annotated extractions. Annotations can be downloaded as a human-readable file or as a JSON file (loadable for further annotation with AnnIE).}
    \label{fig:download}
    \vspace{-1em}
\end{figure}





 
\subsection{Web Interface and Language Support}
%
The user can start from scratch by uploading a text file that contains unannotated sentences, or load previously saved work (JSON file). For each sentence, the user starts from a full set of sentence tokens with highlighted tokens of interest (Figure~\ref{fig:highlight}). The user then constructs triples and places them into fact synsets (Figure~\ref{fig:fact-label}). At any point during the annotation, the user can generate human-readable 
output of the annotated extractions and download it as a text file in a tab-separated format (Figure~\ref{fig:download}). Alternatively, the user can save the annotation progress as a JSON file that can later be loaded in order to continue annotating.

%

AnnIE supports OIE annotations for sentences in any language supported by its NLP module (i.e., available POS-tagging and NER models). By default, AnnIE relies on \texttt{SpaCy} and can therefore support creation of OIE benchmarks for all languages for which \texttt{SpaCy} provides POS-tagging and NER models. 
Section~\ref{app-sec:multilinguality} from the appendix provides details about how 
this module can be adjusted to the user's preference.

%% file: 5-experiments.tex
\section{Demonstration Study}

\setlength{\tabcolsep}{3.5pt}
\begin{table*}[t!]
    \centering
    \def\arraystretch{0.8}
    \footnotesize
        \begin{tabular}{rrrrrrrrrrrrr}  
            \toprule
            & & \multicolumn{5}{c}{\textsc{En}} & \textsc{Zh} & \textsc{De} & \textsc{Ar} & \textsc{Gl} & \textsc{Ja} \\
            \cmidrule(l){3-7}\cmidrule(l){8-8}\cmidrule(l){9-9}\cmidrule(l){10-10}\cmidrule(l){11-11}\cmidrule(l){12-12}
                        &  & ClausIE & MinIE & Stanford   & ROIE & OpenIE6 & M$^2$OIE & M$^2$OIE  & M$^2$OIE & M$^2$OIE & M$^2$OIE \\
            \midrule
            \multirow{2}{*}{P}      & CaRB          & \textbf{0.58}       & 0.45  & 0.17    & 0.44 & 0.48     & /       & /    & / & / & / \\
                                    & Fact-based    & \textbf{0.50}       & 0.43  & 0.11    & 0.20 &    0.31  & 0.18      & 0.09 &    0.16 & 0.15 & 0.00 \\
                                    & $\Delta$ & +0.08 & +0.02 & +0.06 & \textbf{+0.24} & +0.17 & / & / & / & / & / \\
            \midrule
            \multirow{2}{*}{R}      & CaRB          & 0.53       & 0.44  & 0.29    & 0.60 & \textbf{0.67}    & /      & /      &  / & / & / \\                      
                                    & Fact-based    & 0.26       & \textbf{0.28}  & 0.16    & 0.09  & 0.21    & 0.10      & 0.03  &  0.03  & 0.06 & 0.00 \\
                                    & $\Delta$ & +0.27 & +0.16 & +0.13 & \textbf{+0.51} & +0.46 & / & / & / & / & / \\
            \midrule
            \multirow{2}{*}{$F_1$}  & CaRB          & \textbf{0.56}       & 0.44  & 0.22    & 0.51 & \textbf{0.56}    & /      & /    & / & / & / \\
                                    & Fact-based    & \textbf{0.34}       & \textbf{0.34}  & 0.13    & 0.13 & 0.25    & 0.13      & 0.04  & 0.05   & 0.09 & 0.00 \\
                                    & $\Delta$ & +0.22 & +0.10 & +0.09 & \textbf{+0.38} & +0.31 & / & / & / & / & / \\
            \bottomrule
        \end{tabular}
        \vspace{-0.5em}
    
    \caption{Comparison of performance of OIE systems on fact-based v.s.~CaRB benchmarks for English (\textsc{En}), Chinese (\textsc{Zh}), German (\textsc{De}), Arabic (\textsc{Ar}), Galician (\textsc{Gl}) and Japanese (\textsc{Ja}). Metrics used: precision (P), recall (R) and $F_1$ score ($F_1$). $\Delta$ is the difference between the CaRB scores and the fact-based scores.  \textbf{Bold numbers} indicate highest score for English per row (i.e., highest score for P / R / $F_1$ per benchmark) or highes score difference per row (i.e., highest $\Delta$ for P / R / $F_1$ per benchmark). The fact-based benchmark on English reveals that CaRB overestimates the performance of OIE systems, but with the help of AnnIE it is easily possible to create benchmarks which provide more reliable performance estimates. Creating such fact-based benchmarks for a series of other languages highlights the need for future OIE research to focus on langauges other than English.} 
    \label{tab:experiments}
    \vspace{-1em}
\end{table*}

To showcase our tool's suitability for different OIE scenarios, we generated two complete fact-based OIE benchmarks using AnnIE: 
(1) a benchmark for verb-mediated facts; (2) a benchmark with facts involving named entities (NEs). 
We then evaluated several OIE systems and compared their fact-based scores with the token-overlap lenient scores of the existing CaRB benchmark \cite{bhardwaj2019carb}.



\subsection{Experimental Setup}

\textbf{OIE Systems.} We comparatively evaluated several state-of-the-art OIE systems against the gold fact synsets annotated with AnnIE. For OIE on English, we used ClausIE \cite{del2013clausie}, 
Stanford \cite{angeli2015leveraging}, MinIE \cite{gashteovski2017minie}, ROIE \cite{stanovsky2018supervised}
and OpenIE6 \cite{kolluru2020openie6}. 
For Chinese, German, Galician, Japanese and Arabic, we used the supervised M$^2$OIE \cite{ro2020multi2oie} model, which is based on  multilingual BERT \cite{devlin2018bert}, trained on large English dataset \cite{zhan2020span} and transferred to the target language by means of its multilingual encoder. 
We  trained M$^2$OIE 
using the implementation and recommended hyperparameter setup from the original work \cite{ro2020multi2oie}.

\vspace{1.2mm}
\noindent \textbf{Verb-Mediated Triples.} We first evaluate the OIE systems in the most common setup: for verb-mediated facts. In this scenario, OIE system extractions are triples with verb-phrase predicates. We randomly sampled 300 sentences from the CaRB benchmark, and two experts independently annotated them manually for fact synsets with AnnIE (we provide the annotation guidelines in the Appendix, Section~\ref{app:annotation_guidelines_en}). Then, the annotators merged the annotations by resolving the disagreements through a discussion. The annotation effort was approximately two working weeks per annotator. To show that AnnIE is in principle language agnostic, native speakers of German, Chinese, Japanese and Galician translated these 300 sentences to their respective languages. For Arabic, a native speaker managed to translate the first 100 sentences. only due to limited resources. Then, the native speakers annotated  fact synsets in these languages with AnnIE. Due to limited resources, the sentences for these languages were translated and then annotated with OIE extractions by one annotator per language. Finally, we evaluated one OIE system for each language on this benchmark.

\vspace{1.2mm}
\noindent\textbf{NE-centric Triples.} We used AnnIE to build a benchmark consisting of facts connecting named entities (NEs): triples in which both subjects and objects are named entities. 
Since NEs are frequently mentioned in news stories, we selected the sentences for annotation from the NYT10k dataset 
\cite{gashteovski2017minie}, a random sample of 10k sentences from the New York Times corpus \cite{sandhaus2008new}. We split the NE-centric benchmark in two parts: (1) NE-2: 150 sentences from NYT10k with exactly 2 NEs (as detected by \texttt{SpaCy}); (2) 
NE-3$^{+}$: we sample 150 sentences from NYT10k such that they have 3 or more NE mentions. The annotation guidelines, while similar to those for the verb-mediated triples, differ in two important aspects: (1) the annotator should extract \textit{only} the facts in which both arguments are named entities; (2) besides verb-mediated relations, the
annotator was allowed to extract noun-mediated relations too; e.g., \emph{("Sundar Pichai"; "CEO"; "Google")}.


\setlength{\tabcolsep}{2.5pt}
\begin{table}[t!]
    \centering
    \def\arraystretch{0.8}
    \footnotesize
        \begin{tabular}{rlrrrrr}  
            \toprule
                                    &           & \footnotesize{ClausIE}   & \footnotesize{MinIE}     & \footnotesize{Stanford}  &  \footnotesize{ROIE} & \footnotesize{OIE6}  \\
                                    &           & \tiny{(4 / 10)}  & \tiny{\textbf{(26 / 48)}} & \tiny{(12 / 22)} & \tiny{(2 / 7)} & \tiny{(8 / 20)} \\
            \midrule
            \multirow{2}{*}{P}      & NE-2      &   \textbf{0.75}     &   0.58     &   0.45     & 0.05 &   0.38    \\
                                    & NE-3$^{+}$&   \textbf{0.78}     &   0.54     &   0.63     & 0.05 &   0.32    \\
            \midrule
            \multirow{2}{*}{R}      & NE-2      &  0.05      &   \textbf{0.23}     &   0.08     & 0.02 &   0.05     \\
                                    & NE-3$^{+}$&  0.04      &   \textbf{0.13}     &   0.06     & 0.02 &   0.03     \\
            \midrule
            \multirow{2}{*}{$F_1$}  & NE-2      &  0.09      &   \textbf{0.33}     &   0.13     & 0.03 &   0.08     \\
                                    & NE-3$^{+}$&  0.07      &   \textbf{0.21}     &   0.11     & 0.02 &  0.06     \\
            \bottomrule
        \end{tabular}
        \vspace{-0.5em}
    
    \caption{Performance of OIE systems on fact-based evaluation on NE-centric triples. NE-2 / NE-3$^{+}$: results on sentences that contain 
    2 / 3 or more NEs (labelled with SpaCy). Numbers in brackets below an OIE system name indicate
    the number of OIE triples on which the evaluation was done for NE-2 / NE-3$^{+}$. \textbf{Bold numbers} indicate highest score per row.} 
    \label{tab:experiments-ner}
    \vspace{-1em}
\end{table}

\subsection{Results and Discussion}

\vspace{1.2mm}
\noindent\textbf{English OIE.}
We score the OIE systems against the gold fact synsets produced with AnnIE, using the fact-based evaluation protocol 
(Section~\ref{sec:benchie}). For the verb-mediated extractions, we compare our fact-based evaluation scores against the token-overlap scores of CaRB \cite{bhardwaj2019carb}: the results are shown in Table~\ref{tab:experiments}. Comparison of Fact-based and CaRB scores indicates that:~(1) CaRB largely overestimates the performance of OIE systems; (2) current OIE systems predominantly fail to extract correct facts,  which strongly points to the need for creating complete fact-oriented benchmarks, a task that AnnIE facilitates. For more detailed discussion, multi-faceted evaluation and error analysis, see \citet{gashteovski2021benchie}.

\vspace{1.2mm}
\noindent\textbf{Multilingual OIE.}
Finally, Table~\ref{tab:experiments} shows that the results for OIE systems in languages other than English are significantly worse, which shows that more research is needed in this direction.
The results for Chinese OIE seem to be particularly encouraging, as the difference of the $F_1$ score between some OIE systems in English and M$^2$OIE in Chinese is not too large as it is between English and the other languages. For example, M$^2$OIE in Chinese has the same $F_1$ score as Stanford's OIE system and ROIE. We applied the same training strategy for Japanese, but the $F_1$ score of this OIE system is 0. This indicates that more research is needed for Japanese in defining the problem well and proposing methods for solving it. Nevertheless, we release the gold datasets for OIE in all investigated languages: English, German, Galician, Chinese, Japanese and Arabic; which we believe to be an important resource for research for subsequent multilingual OIE. Figure \ref{fig:annie-multiling1} and Figure \ref{fig:annie-multiling2} from the Appendix show an example sentence and its corresponding OIE annotations in different languages. For a more detailed discussion on the results of  multilingual OIE evaluation, see \citet{kotnis2022integrating}.

\vspace{1.2mm}
\noindent\textbf{NE-centric OIE.}
Table~\ref{tab:experiments-ner} shows the performance of OIE systems on the NE-centric benchmark. In both subsets of 150 sentences---NE-2 and NE-3$^{+}$---only a fraction of them contain actual knowledge facts that connect a pair of NEs (59/150 and 97/150 respectively). Because the OIE systems used in the evaluation are not specifically designed to extract NE-centric facts, we make the evaluation fairer by pruning the system extractions before fact-based evaluation: we keep only the triples that contain in both subject and object NEs found among subjects and objects of gold extractions. In other words, we primarily test whether the OIE systems extract acceptable predicates between NEs between which there is a predicate in the gold standard. The results show that the current OIE systems extract very few NE-centric triples (e.g., ClausIE extracts only 4 triples for the NE-2 
dataset and 10 for the NE-3$^{+}$ dataset, see Table~\ref{tab:experiments-ner}). Because of this, one should intepret the results in Table~\ref{tab:experiments-ner} with caution. 
This evaluation, nonetheless shows that the current 
OIE systems are not ill-suited for a NE-centric OIE task, warranting more research efforts in this direction.

%% file: 6-conclusions.tex
\section{Conclusions}
Exhaustively annotating all acceptable OIE triples is a tedious task, but important for realistic intrinsic evaluation of OIE systems. To support annotators, we introduced AnnIE: annotation tool for constructing comprehensive evaluation benchmarks for OIE. AnnIE allows custom specification of tokens of interests (e.g., verbs) and is designed for creating fact-oriented benchmarks in which the fact-equivalent--yet superficially differing--extractions are grouped into fact synset. 
AnnIE's lightweight architecture, easy installation and customizable components make it a practical solution for future OIE annotation.

\section*{Acknowledgments}

We thank the anonymous reviewers for their invaluable feedback and support.

%% file: appendix.tex
\section{Appendix}

\subsection{Highlighting Functions}
\label{app:highlighting-functions}

\begin{figure}
    \includegraphics[scale=0.5]{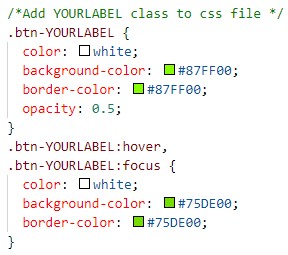}
    \caption{Add new class to css}
    \label{fig:adjust-css}
\end{figure}

The tool is designed to make customizations rather easily. For adjusting the color scheme, the color “hex”-values inside the style.css files need to be set to the desired color codes. This needs to be done to the button itself with the desired color as well as to the “hover” property of the button, where usually a darker version of the same color is used. If complete new labels are introduced to the tool, the css needs to include an approriate class to handle these as shown in Figure~\ref{fig:adjust-css}.

\begin{figure*}
    \includegraphics[scale=0.55]{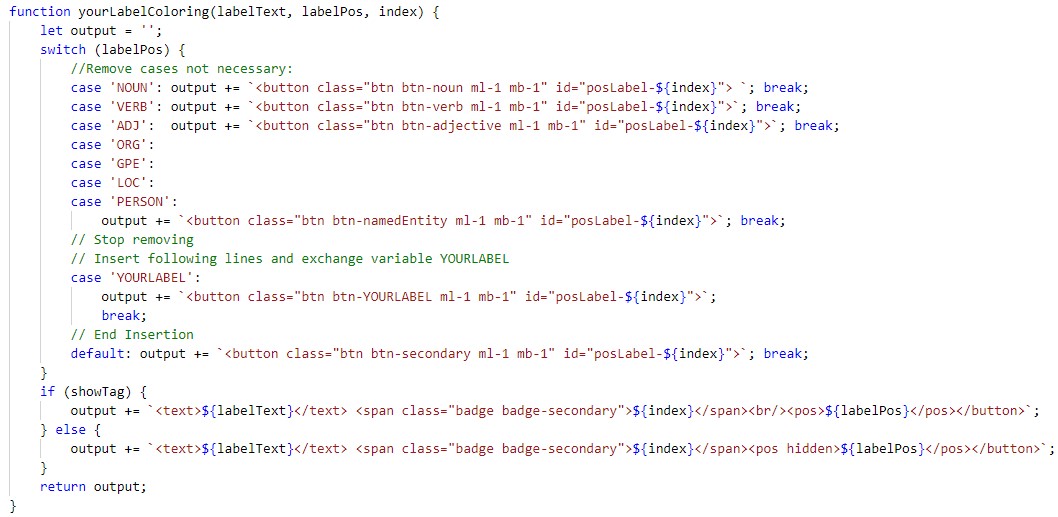}
    \caption{Example of a new coloring function}
    \label{fig:adjust-yourFct}
\end{figure*}

\begin{figure*}
    \includegraphics[scale=0.55]{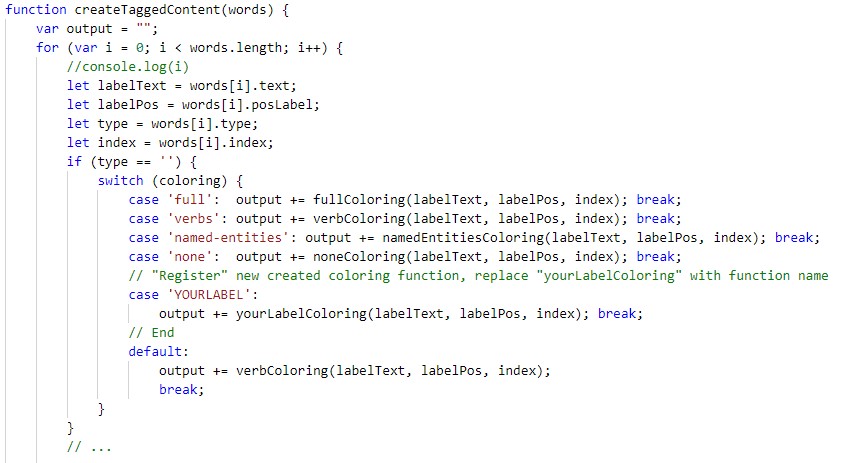}
    \caption{"Register" new coloring function}
    \label{fig:adjust-register}
\end{figure*}

In case any new coloring schemes are required, these can either be entered additionally or exchanged against the standard functions implementing the scheme above. The different colorings are applied using the functions \texttt{fullColoring()}, \texttt{verbColoring()}, \texttt{namedEntitiesColoring()}, and \texttt{noneColoring()} inside \texttt{GraphicalInterface.js}. These functions can be adjusted by changing the switch statements handling which tokens need to be colorized depending on their label. There, new cases can simply be added or superfluous colored labels can be removed.Another option is to add new coloring functions. These can rely on the provided ones. They simply need to be registered to the tool by being added to the first switch statement inside the function \texttt{createTaggedContent()}. An example of such an function is given in Figure~\ref{fig:adjust-yourFct}, while the "register" procedure is shown in Figure~\ref{fig:adjust-register}.

\begin{figure*}
    \includegraphics[scale=0.6]{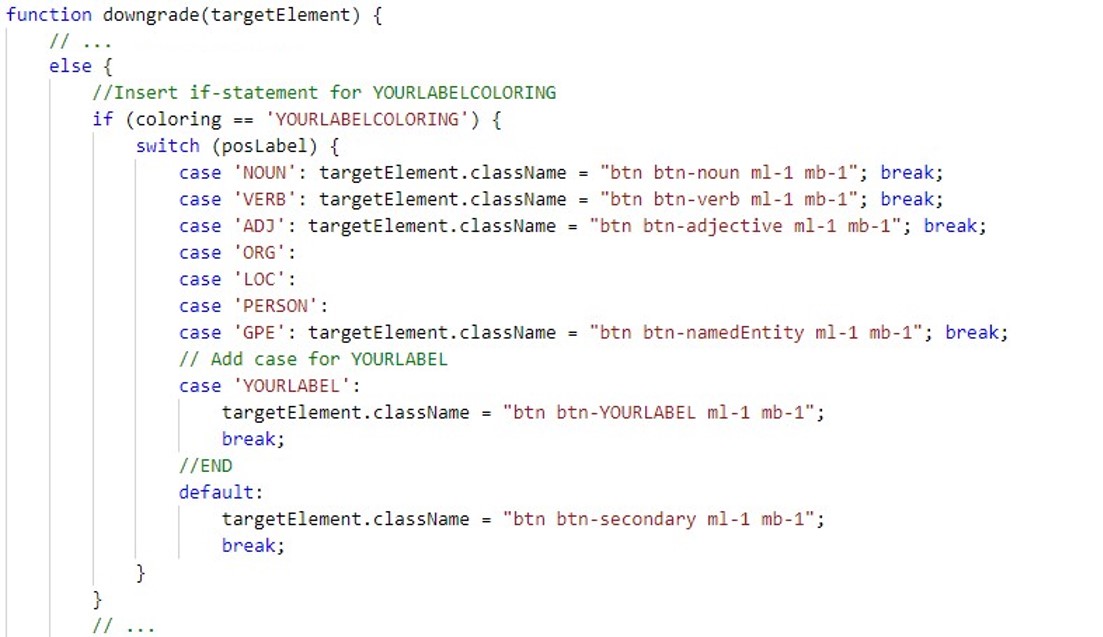}
    \caption{Necessary adjustments to \texttt{downgrade()}}
    \label{fig:adjust-downgrade}
\end{figure*}

In both cases, additionally, the function \texttt{downgrade()} needs to be adjusted accordingly to the above-mentioned changes to ensure that the buttons can be selected and deselected properly. This step is shown in Figure~\ref{fig:adjust-downgrade}.

\subsection{Data model structure used in the frontend}
\label{app-subsec:data-model}

\begin{figure}
    \includegraphics[scale=0.6]{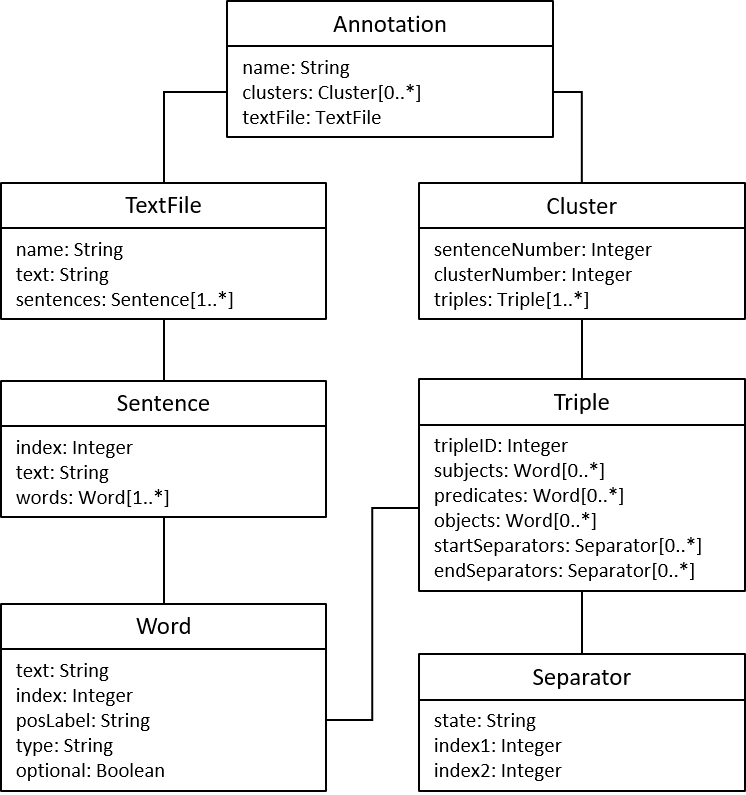}
    \caption{Data model structure used in the frontend}
    \label{fig:model_structure}
\end{figure}

The data model structure used in the frontend is shown on Figure~\ref{fig:model_structure}.

\subsection{Multilinguality}
\label{app-sec:multilinguality}
\begin{figure}
    \includegraphics[scale=0.52]{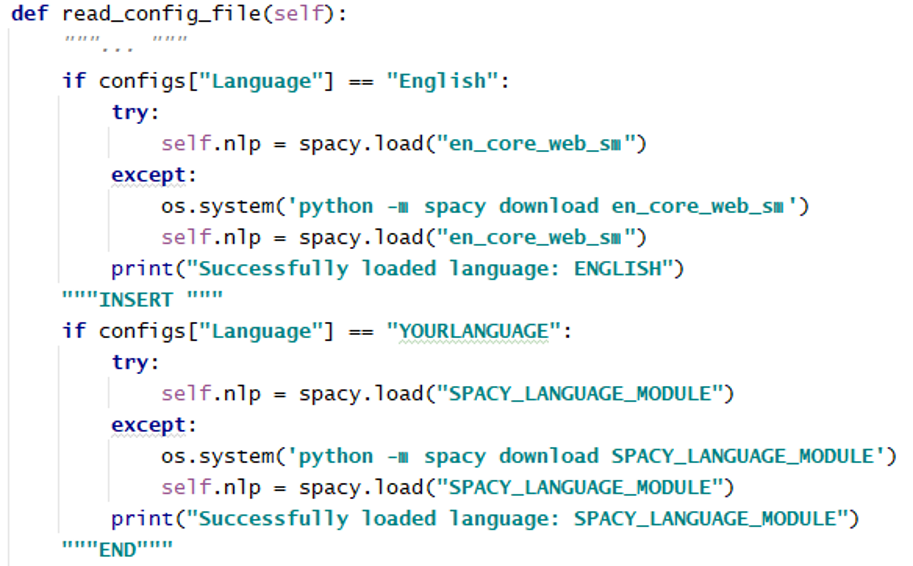}
    \caption{Code snippet showing how to add further language support}
    \label{fig:additional-languages}
\end{figure}

By extending the definition of the function \texttt{read\_config\_file()} inside the backend file \texttt{tokenizer.py}, further languages can be included. Therefore, SpaCy simply needs to be forced to load the appropriate language model. For details, see code snippet showing how to adapt the \texttt{tokenizer.py} script to accept further language models in Figure~\ref{fig:additional-languages}.

\subsection{Annotation Guidelines (English)}
\label{app:annotation_guidelines_en}

\subsubsection{General Principle} 
The annotator should manually extract verb-mediated triples from a natural language sentence. Each triple should represent 
two entities or concepts, and the verb-mediated relation between them. For example, from the input sentence \emph{"Michael Jordan, who is a former basketball player, was born in Brooklyn."}, there are three entities and concepts---\emph{Michael Jordan, former basketball player} and \emph{Brooklyn}---which are 
related as follows: \emph{("Michael Jordan"; "is"; "former basketball player")} and \emph{("Michael Jordan"; "was born in"; "Brooklyn")}.

Once the triple is manually extracted, it should be placed into the correct fact synset (see Section~\ref{app:fact-synset}).

\subsubsection{Fact Synsets}\label{app:fact-synset}
Once a triple is manually extracted, the annotator should place the triple into its corresponding fact synset (details about 
fact synsets in Section~\ref{sec:benchie}). In case there is no existing fact synset for the manually extracted triple, the annotator should create 
one and place the triple in that synset. 

\textbf{Coreference.} The annotator should place extractions that refer to the same entity or concept under the same fact synset. Consider the following
input sentence: \emph{"His son , John Crozie, was an aviation pioneer."}; The following triples should be placed in the same fact synset:
\begin{itemize}
    \item \emph{("His son"; "was"; "[an]\footnote{words in square brackets indicate optional tokens (see Section \ref{app:subsec-opt-tokens})} aviation pioneer")}
    \item \emph{("J.~Crozie"; "was"; "[an] aviation pioneer")}
\end{itemize}
because \emph{"His son"} and \emph{"John Crozie"} refer to the same entity.

\textbf{Token placements.} The annotator should consider placing certain tokens in different slots, without damaging the meaning of each the fact. 
Consider the input sentence \emph{"Michael Jordan was born in Brooklyn."}. There is one fact synset and its corresponding triples:
\begin{enumerate}
    \item[$f_1$] \emph{("M.~J."; "was born in"; "Brooklyn")} \\
                 \emph{("M.~J."; "was born"; "in Brooklyn")}
\end{enumerate}
In the first triple, the preposition \emph{"in"} is in the relation, while in the second it is in the object. The annotator should allow for 
such variations, because OIE systems should not be penalized for placing such words in different slots.

\subsubsection{Optional Tokens} 
\label{app:subsec-opt-tokens}
If possible, the annotator should label as \emph{optional} all tokens that can be omitted in an extraction without damaging its semantics.
Such tokens include determiners (e.g., \emph{a, the, an}), honorifics (e.g., \emph{[Prof.]~Michael Jordan}) or certain quantities (e.g., 
\emph{[some] major projects}. The optional tokens are marked with square brackets $[~]$. 

In what follows, we show examples of considered optional token(s).

\textbf{Determiners.} Unless a determiner is a part of a named entity (e.g., \emph{"The Times"}), it is considered as optional. For instance, the following
triples are considered to be semantically equivalent: 
\begin{itemize}
    \item \emph{("Michael Jordan"; "took"; "the ball")}
    \item \emph{("Michael Jordan"; "took"; "ball")}
\end{itemize}
The annotator, therefore, should annotate \emph{("Michael Jordan"; "took"; "[the] ball")}, where the optional token is in square brackets.

\textbf{Titles.} Titles of people are considered optional; e.g., \emph{("[Prof.] Michael Jordan"; "lives in"; "USA")}.

\textbf{Adjectives.} The annotator should label adjectives as optional if possible. For example, in the following triple, the adjective 
\emph{outstanding} can be considered optional: \emph{("Albert Einstein"; "is"; "[an] [outstanding] scientist")}. Note that the annotator should be 
careful not to label adjectives as optional if they are essential to the meaning of the triple. For instance, the adjective \emph{cold} should not 
be labeled as optional in the triple \emph{("Berlin Wall"; "is infamous symbol of"; "[the] cold war")}. 

\textbf{Quantities.} Certain quantities that modify a noun phrase can be considered as optional; e.g., \emph{("Mitsubishi"; "has control of"; 
"[some] major projects")}.

\textbf{Words indicating some tenses.} The annotator can treat certain verbs that indicate tense as optional. For instance, the word \emph{have} in 
\emph{("FDA"; "[has] approved"; "Proleukin")} can be considered as optional, since both VPs \emph{"have approved"} and \emph{"approved"} contain 
the same core meaning. 

\textbf{Verb phrases.} It is allowed for the annotator to mark verb phrases as optional if possible; e.g. \emph{("John"; "[continues to] reside in";
"Berlin")}.

\textbf{Passive voice.} When possible, if an extraction is in passive voice, the annotator should place its active voice equivalent 
into the appropriate fact synset. For instance, suppose we have the sentence \emph{"The ball was kicked by John.".} Then, the fact synset should 
contain the following triples:
\begin{itemize}
    \item \emph{("[The] ball"; "was kicked by"; "John")}
    \item \emph{("John"; "kicked"; "[The] ball")}
\end{itemize}
Note that the opposite direction is not allowed. If the sentence was \emph{"John kicked the ball."}, then the annotator is not allowed to manually 
extract the triple \emph{("[The] ball"; "was kicked by"; "John")} because such extraction contains words that are not originally found in the input sentence 
(\emph{"was"} and \emph{"by"}). These are so-called implicit extractions and we do not consider them (details in Sec.~\ref{app:subsec-implicit})).

\textbf{Attribution clauses.} Extractions that indicate attribution of information should be placed in the same fact synset as the original information 
statement. For example, the core information of the sentence \emph{"Conspiracy theorists say that Barack Obama was born in Kenya."} is that Obama was 
born in Kenya. As indicated by \citet{mausam2012open}, it is important not to penalize OIE systems that would also extract the context about the 
attribution of such information. Therefore, the annotator should include the following triples into the same fact synset: \emph{("Barack Obama"; 
"was born in"; "Kenya")} and \emph{("Conspiracy theorists"; "say that"; "Barack Obama was born in Kenya")}. 

\subsubsection{Incomplete Clauses} 
The annotator should not manually extract incomplete clauses, i.e., triples such that they lack crucial piece of information. Suppose there is the 
input sentence \emph{"He was honored
by the river being named after him".} The following triple should not be manually extracted: \emph{("He"; "was honored by"; "[the] river")}, but the 
following triples should be: \emph{("He"; "was honored by [the] river being named after"; "him")} and \emph{("[the] river"; "being named after"; "him")}.

\subsubsection{Overly Complex Extractions}
The annotators should not manually extract overly specific triples, such that their arguments are complex clauses. For instance, for the input sentence 
\emph{"Vaccinations against other viral diseases followed, including the successful rabies vaccination by Louis Pasteur in 1886."}, the following triple
should not be extracted: \emph{("Vaccinations against other viral diseases"; "followed"; "including the successful rabies vaccination by Louis Pasteur in 1886")} because the object is a complex clause which does not describe a single concept precisely, but rather it is composed of several concepts.

\subsubsection{Conjunctions}
The annotator should not allow for conjunctive phrases to form an argument (i.e., subject or object). Such arguments should be placed into separate 
extractions (and in separate fact synsets). Consider the sentence \emph{"Michael Jordan and Scottie Pippen played for Chicago Bulls.".} The annotator should manually extract the 
following triples:
\begin{itemize}
    \item \emph{("M.~Jordan"; "played for"; "Chicago Bulls")}
    \item \emph{("S.~Pippen"; "played for"; "Chicago Bulls")}
\end{itemize}
The annotator should not, however, manually extract \emph{("Michael Jordan and Scottie Pippen"; "played for"; "Chicago Bulls")}.

\subsubsection{Implicit Extractions} 
\label{app:subsec-implicit}
We focus on explicit extractions, which means that every word in the extracted triple must be present in the original input sentence. Therefore, implicit extractions---i.e., extractions that contain inferred information which is not found in the sentence explicitly---are 
not considered. 
One example implicit extraction is \emph{("Michael Jordan"; "be"; "Prof.")} from the input sentence \emph{"Prof.~Michael Jordan lives in
USA."}, where the triple infers that Michael Jordan is professor without being explicitly indicated in the sentence (i.e., the word \emph{"be"} is not present 
in the input sentence, it is inferred).

\subsection{Annotation Guidelines (Chinese)}
\label{app:annotation_guidelines_zh}

The annotator followed the same general principles as with the English annotation guidelines (Sec.~\ref{app:annotation_guidelines_en}. 
Due to the difference of the languages, we slightly adapted the annotation guidelines for the Chinese language. In what follows, we list those differences.

\subsubsection{Articles}
Chinese language does not contain articles (i.e., \emph{"a", "an", "the")}. Therefore, in the manual translation of the sentences, there are no 
articles in the Chinese counterparts, which also results in labeling such words as optional (for English, see Sec.~\ref{app:subsec-opt-tokens}).

\subsubsection{Prepositional Phrases within a Noun Phrase}
Certain noun phrases with nested prepositional phrase cannot be translated directly into Chinese the same way as in English. For example, suppose we have
the phrase \emph{"Prime Minister of Australia"}. In Chinese, the literal translation of this phrase would be \emph{"Australia's Prime Minister"}. For instance,  
in the English annotations the sentence \emph{"He was the Prime Minister of Australia"} would have two fact synsets:
\begin{itemize}
    \item[$f_1$] \emph{("He"; "was [the] Pr.~Min.~of"; "Australia")}
    \item[$f_2$] \emph{("He"; "was"; "[the] Pr.~Min.~[of Australia]")}
\end{itemize}
This is because the the fact synset $f_1$ relates the concepts \emph{"he"} and \emph{"Australia"} with the relation \emph{"was [the] Prime Minister of"}, while the second 
fact synset relates the concepts \emph{"he"} and \emph{"Prime Minister [of Australia]"} with the relation \emph{"was"}. 

In Chinese language, however, the construction of $f_1$ would not be possible, because the phrase \emph{"Prime Mininister of Australia"} 
cannot be separated into \emph{"Prime Minister"} and \emph{"Australia"}. Therefore, the golden annotation for this particular example in Chinese would 
be only one fact synset: \emph{("He"; "was"; "[Australia's] Prime Minister")}, which is equivalent with $f_2$. 


\subsection{Manual Labeling of OIE Triples}
\label{app:gui}

\begin{figure*}[h]
    \centering
    \includegraphics[scale=0.75]{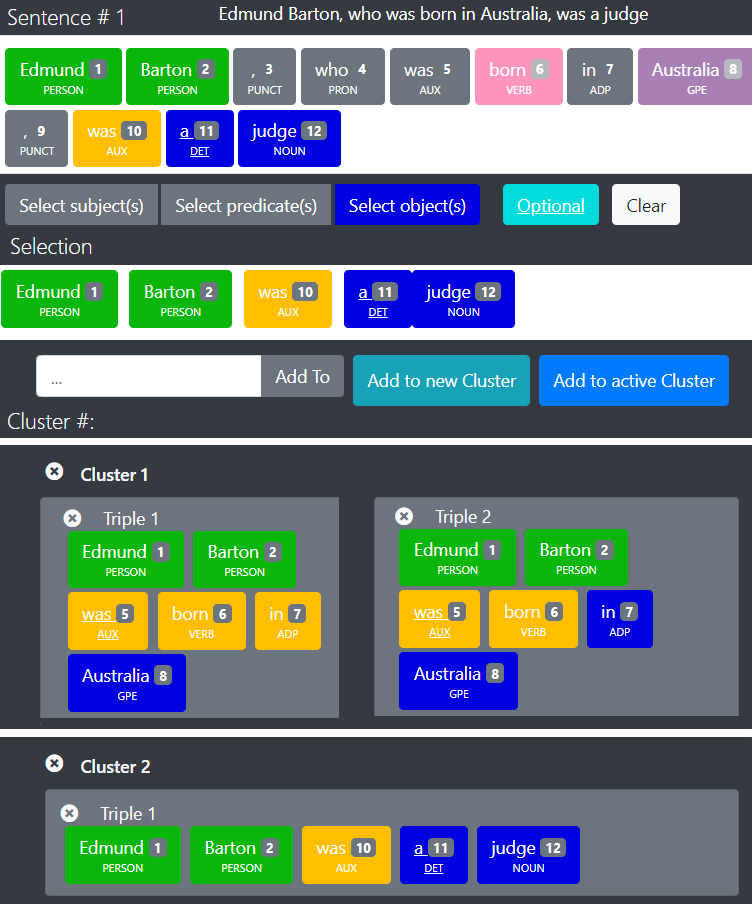}
    \caption{Manual labeling of OIE triples. The user selects tokens from the tokenized input sentence and 
    places them into the correct slot: \textcolor{forestgreen}{subject (green)}, \textcolor{orange}{predicate (yellow)} or \textcolor{blue}{object (blue)}. Then, the user adds the extracted triple either to an active fact cluster (i.e., fact synset) or to a new one. The user can also select which tokens are optional by clicking the "Optional" button on an active token selection.}
    \label{fig:fact-label-large}
    \vspace{-1em}
\end{figure*}

See Figure \ref{fig:fact-label-large} for a screenshot from AnnIE's GUI, which shows the manual labeling process of OIE triples given an input sentence. 

\begin{figure*}
    \centering
    \includegraphics[scale=0.85,trim={4.5cm 8.5cm 4.5cm 0cm}]{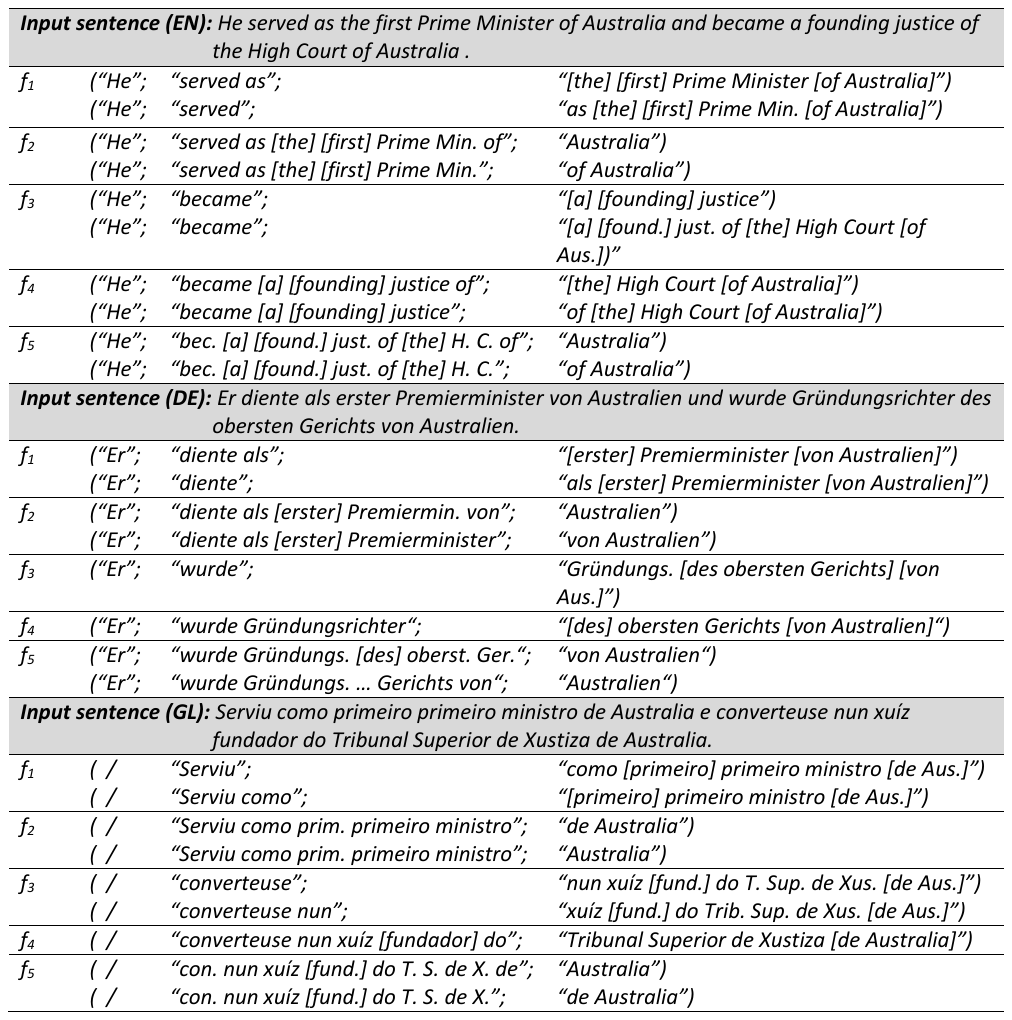}
    \caption{Example sentence with five \textit{fact synsets} ($f_1$--$f_5$) in several languages: English (EN), German (DE) and Galician (Galician). Square brackets ([]) are a shorthand notation for multiple extractions: triples both with and without the expression(s) in the brackets are considered correct. For continuation of this figure, see Figure \ref{fig:annie-multiling2}, whereas the same input sentence and its corresponding OIE annotations are written in Chinese (ZH), Japanese (JA) and Arabic (AR).}
    \label{fig:annie-multiling1}
    \vspace{-1em}
\end{figure*}

\begin{figure*}
    \centering
    \includegraphics[scale=0.85,trim={4.5cm 11cm 4.5cm 0.5cm}]{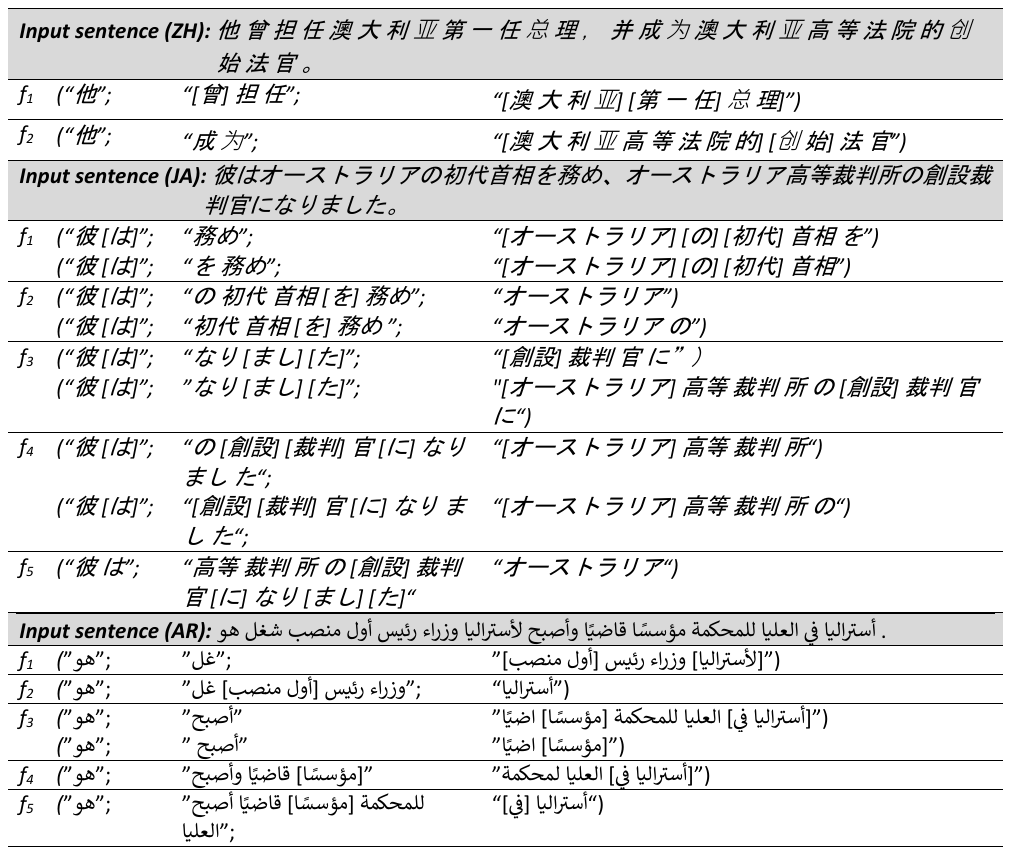}
    \caption{Example sentence and its corresponding OIE annotations in Chinese (ZH), Japanese (JA) and Arabic (AR). This figure is continuation of Figure \ref{fig:annie-multiling1}.}
    \label{fig:annie-multiling2}
    \vspace{-1em}
\end{figure*}
